# Segmenting Bank Customers via RFM Model and Unsupervised Machine Learning

Musadig Aliyev
School of Information Technologies and Engineering
ADA University
Baku, Azerbaijan
maliyev2019@ada.edu.az

Elvin Ahmadov
School of Information Technologies and Engineering
ADA University
Baku, Azerbaijan
eahmadov2019@ada.edu.az

Habil Gadirli
School of Information Technologies and Engineering
ADA University
Baku, Azerbaijan
hgadirli2019@ada.edu.az

Arzu Mammadova
School of Information Technologies and Engineering
ADA University
Baku, Azerbaijan
amammadova2019@ada.edu.az

Emin Alasgarov
School of Information Technologies and Engineering
ADA University
Baku, Azerbaijan
ealasgarov@ada.edu.az

*Abstract*—In recent years, one of the major challenges for financial institutions is the retention of their customers using new methodologies of reliable and profitable segmentation. In the field of banking, the approach of offering all of the services to all the existing customers at the same time does not always work. However, being aware of what to sell, when to sell and whom to sell makes a huge difference in the conversion rate of the customers responding to new services and buying new products. In this paper, we used RFM technique and various clustering algorithms applied to the real customer data of one of the largest private banks of Azerbaijan.

## I. Introduction

Nowadays, one of the major objectives in the industries, especially in the banking sector is to understand customers' behaviour and classify them into appropriate groups based on the obtained result. Companies are looking for answers to the questions, such as who are the most preferred customers, who are loyal to the company, which product may attract more customers on the shelves of a retail outlet to increase the sales, etc.

Consequently, it can be emphasized that the problem of this project is a customer segmentation, which allows businesses to gain more insight about customers' behavior in order to satisfy their needs more efficiently. Accordingly, the purpose of the project was to evaluate the association of unsupervised machine learning algorithms in combination with RFM (recency, frequency and monetary value) model to find the optimal number of customer segments. In appliance with that, we have used a real customer data that is available at Unibank (one of the leading retail banks in Azerbaijan) to divide customers into clusters (i.e. segments) based on proposed methodologies.

## II. Literature Review

To begin, we have reviewed several papers related to the topic of customer retention, customer segmentation and personalized offers. Initially, we reviewed a paper named "Classifying the segmentation of customer value via RFM model and RS theory", done by Ching-Hsue Cheng, You-Shyang Chen [1]. This paper suggests a procedure not only to increase and improve classification accuracy but also to derive out the classification rules to have an excellent customer management. Furthermore, while reading this paper we have seen the drawbacks of data mining tools and the ways of developing them. Authors also propose the methods and techniques to easily and objectively group the customer segmentation.

Another paper called "K-modes Clustering Algorithm for Categorical Data" by N. Sharma and N. Gaud [2], gave us sufficient information regarding the formulas and mathematical approaches of K-modes algorithm. Moreover, it explains K-modes being an extension to the standard K-Means clustering algorithm and shows the main modifications to K-Means as well. One more important material that we looked through regarding K-Means technique was named "Implementing & Improvisation of K-Means Clustering Algorithm" by Unnati R. Raval and Chaita Jani [3]. In this work, the authors highlight the clustering techniques as the most important part of the data analysis and refer to K-Means as one of the oldest and popular clustering techniques. Also, we found out the advantages and disadvantages of K-Means algorithm and the ways along with techniques to improve the existing algorithm for better accuracy and performance.

Shreya Tripathi, Aditya Bhardwaj and Poovammal E. in their paper called "Approaches to Clustering in Customer Segmentation" [4], examine some clustering algorithms including K-Means and Hierarchical Clustering from various sides, and indicate the final results of comparing these techniques. Also, it shows a perfect correspondence in time, place and circumstance for using each of the algorithms. During the research process, the authors pointed out the advantages and disadvantages of these clustering algorithms.

According to the paper "Revised DBSCAN algorithm to cluster data with dense adjacent clusters" [5], the algorithm has been widely used in many areas of science because of its plain structure and the ability to detect clusters of various sizes. With the help of DBSCAN algorithm, the tight areas are found and recursively expanded to find out dense arbitrarily shaped clusters. However, when detecting border objects of adjacent clusters, the algorithm can be unbalanced. The ultimate clustering result obtained from DBSCAN depends on the order in which objects are processed during the algorithm run.

"Developing a model for measuring customer loyalty and value with RFM technique and clustering algorithms" [6] is another work that we have reviewed. Nowadays, the customer relationship management strategies play a considerable role in business areas. According to this paper, knowing their customers behavior, priorities, and needs based on data allows banking sector to have a specific group of customers to suggest them personalized offers and increase their usage time of existing products. Also, the authors describe RFM technique and different clustering algorithms for customer segmentation. Ultimately, the concept of customer loyalty and retention through behavioral and demographic features was considered one of the important aspects in the paper.

It can be concluded that by analyzing these research papers we made a benchmark of existing solutions on the market along with various techniques and algorithms applied with respect to the problem of customer segmentation. In our project we have decided to utilize the power of certain clustering algorithms in combination with popular RFM technique used for this problem domain.

III. RESEARCH METHODOLOGY

Our research methodology was divided into two directions, which are behavior based segmentation and demographic based segmentation.

The first stage was done through the *behavioral segmentation*, which groups customers based on some similarities that include specific behavior or attitude like consumption, usage and profit toward some products while purchasing them. During our project, three clustering algorithms of different types were used, which are *K-means, Hierarchical* and *DBSCAN*. Moreover, the bank customer features including *Card_Holder, First_Name, Surname, Transaction_ID, Amount, OP_Date_Precise* were used in our model. Then, these features were used in the analysis and preprocessing stages to get the final features for building our behavioral model.

The second stage was performed through the *demographic segmentation*. During this segmentation, different demographic variables like age, gender, region, and income were used, which helped to divide number of people into specific customer groups. To be more specific, K-Modes algorithm was used for demographic model; however, the obtained result was not satisfactory and therefore, the decision was to concentrate on behavioral model for the rest of this project.

As the given data was real and it was related to real bank customers, the quality was not good enough to work on it. Exploratory Data Analysis approach was used to find correlations between features, outliers in the dataset and also, summarize statistics about the dataset and visualize data. Figure 1 shows "bird-view" of the steps taken for the realization of this project.

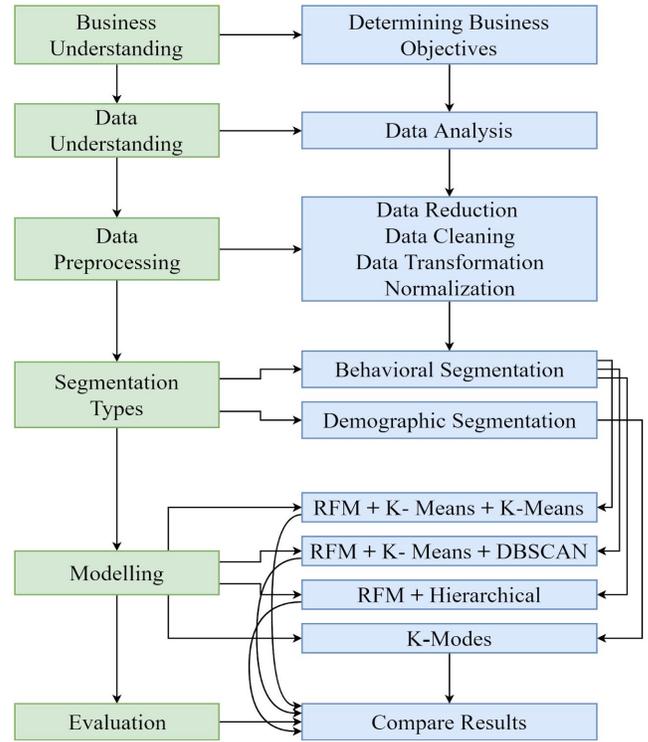

Fig. 1. General overview

Three important steps should be stated explicitly, namely data preprocessing, RFM approach, and RFM Analysis.

*A. Data Preprocessing*

Data preprocessing stage can be named as one of the challenging and hard stages of our project. Due to the reason that distinct transaction ID was used to extract data from the database, there appeared data duplicates, which were dropped afterwards. Because of low correlation some features were removed and the missing values were filled with mean value of corresponding features. Among many normalization methods standardization was selected to be performed on each of the features. The reasons behind this decision were eliminating the bias as much as possible, preserving the scale factor between data points and hence, keeping the outliers in the dataset.

*B. RFM Approach*

To specify, one of the proven marketing techniques that is used to calculate customers' value based on their transaction/purchase history is RFM model. Therefore, to use the modern approach and also to get the desired result, in this project RFM model was chosen to calculate the customers' value for the financial institution. Afterwards, with the aim of calculating 3 features of the RFM model, first dataset was reduced to 4 features, which are: Card_Holder, Transaction ID, Amount, and Date. Then using groupby operation on Card_Holder these 4 existing features were transformed to the RFM features which are Recency, Frequency and Monetary. Hence, RFM values were calculated to complete this step.

Below, the technique for calculating the RFM features is given:

**Recency of a customer:**
d : The most recent date in the table + 1 day
c : The date of the most recent transaction of the customer

$$recency(c) = d - c \qquad (1)$$

**Frequency of a customer:**
In order to find the frequency of the customer the number of transactions (count) made by this customer should be calculated.
$T_c$ : is the set containing transactions made by a particular customer.
id : is the id of the transaction in a given set

$$frequency(c) = \sum_{id \, \varepsilon \, T_c} 1 \qquad (2)$$

**Monetary value of a customer:**
To find the monetary of the customer the amount spent in each transaction should be summed up.

amount(id) : function returns the amount spent for the transaction identified by its id.

$$monetary(c) = \sum_{id \, \varepsilon \, T_c} amount(id) \qquad (3)$$

*C. RFM Analysis*

It is important to note that one of the most widely used customer segmentation techniques is RFM analysis which assigns customers a score based on their calculated RFM values.

To check its precision a model was built based on the RFM analysis technique using Unibank's data. However, after analysing the results obtained, it has been concluded that the model based on the RFM scores does not provide reliable information. To prove this statement let's analyze the model.

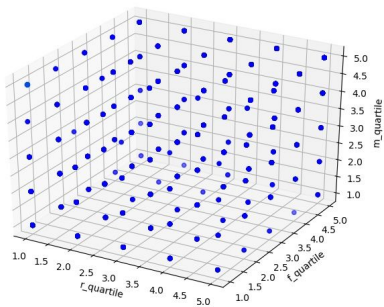

Fig. 2. Clusters obtained by applying RFM analysis technique

In this model each feature of the RFM score was calculated based on the 5 quantiles, meaning there can be maximum score of 1 and a minimum score of 5 for a particular feature. Since there are 3 features and 5 possible values for each of these features, there are 125 possible RFM scores/clusters in total.

$$Total\ possible\ number\ of\ RFM\ scores\ =\ x^3 \qquad (4)$$

x: the number of quantiles a feature is divided into.

Thus, it does not matter whether dataset contains thousand or billion entries, the graph shown in Figure 2 will always be the same (assuming that all (125) of the RFM scores will be assigned to at least one of the customers). This means this technique will always roughly estimate the clusters of the customers.

Considering these limitations of RFM analysis, it has been decided not to use this technique to build a model. Instead RFM values were used with unsupervised machine learning algorithms to build 3 powerful models. As described in Figure 1, these 3 models are based on the 3 different types of algorithms which are: K-Means (centroid based), DBSCAN (density based) and Agglomerative clustering (Hierarchical clustering based).

1. 1st Model is built by using K-means twice.
2. 2nd Model was built by using synthesis of DBSCAN and K-means.
3. 3rd Model was built by using Hierarchical clustering.

IV. IMPLEMENTATION

In this section, implementation of 3 behavioral models introduced in the research methodology section are discussed in detail.

*A. 1st Model(K-means ➔ K-Means)*

K-Means is a centroid based clustering algorithm, which calculates the predetermined number of clusters based on the euclidean distance between data points and centroids. In order to determine the number of clusters "Elbow Method" is used. According to this method model should be built several times by assigning different number of centroids (clusters) and for each of the model WCSS (within cluster sum of errors) should be calculated. After building the models and calculating the WCSS, the graph which depicts WCSS for each model is plotted. The decision about the number of clusters is given based on the slope of the graph.

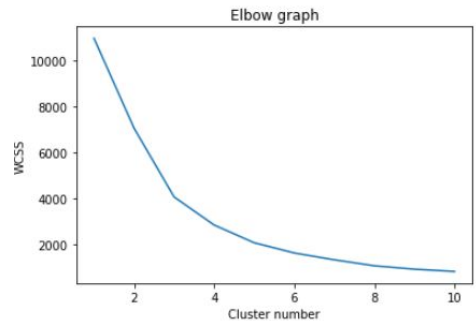

Fig. 3. WCSS for each model based on the K-Means with different number of clusters (centroids)

As is shown in Figure 3, there can be chosen 3, 4, or 5 logical clusters based on this graph. But in this project 4 clusters has been chosen for the model based on the K-means algorithm, since the value of the slope of the graph after 4 (Cluster number) is relatively small.

In order to see the calculated clusters more precisely and to get more insights about them, 3D graph depicting all 4 clusters, each with distinct color, based on Recency, Frequency, and Monetary were plotted.

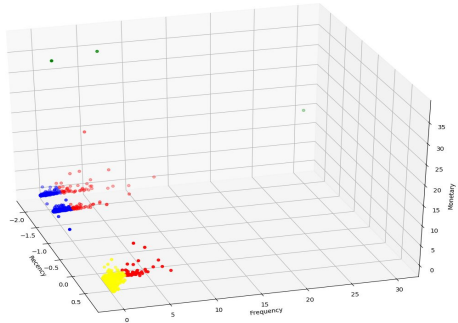

Fig. 4. Clusters calculated by K-means algorithm

After some observations it has been decided to apply K-Means to the red cluster in order to separate it into 2 clusters. The reason behind this decision is that red cluster could be much more valuable if the side with low recency (left side of the cluster as shown in Figure 4) were separated from the other side (right side of the cluster) with high recency to identify active and inactive customers separately.

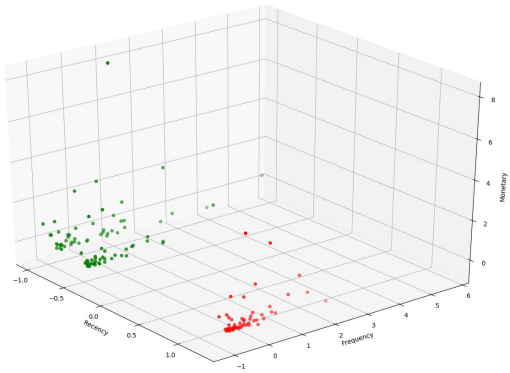

Fig. 5. Cluster obtained by applying K-Means on a red cluster

As it can be seen from Figure 5, K-Means grouped customers with low recency into a green cluster and with high recency into a red cluster. After dividing red cluster into 2 separate clusters the result was merged with the clusters which were obtained during the 1st iteration, with the aim of bringing together all of the obtained clusters by using this model.

### B. 2nd Model (DBSCAN ➔ K-Means)

DBSCAN (Density-based spatial clustering of applications with noise) algorithm as its name indicates, calculates clusters based on the density in the dataset. DBSCAN does not require the predetermined number of clusters as an argument, when building a model based on it, but rather it requires e (points within a radius 'e' of a point is considered as its neighbours) and 'minPoints' (minimum number of points required for a group of points to be considered as a cluster) arguments. Determining values for these arguments depends on the dataset, therefore finding the best values requires some repeating steps. The algorithm for finding these values is given below:

1. Try different values for e and minPoints.
2. Build model based on these values.
3. Analyze the resulting clusters.
4. Decide if the model expresses the dataset reasonably.
   - If yes : use these values.
   - Else : go back to step 1.

In this project 'e' and 'minPoints' has been decided to be 0.8 and 5 respectively, by applying the algorithm above. Thus the model based on DBSCAN algorithm was built based on these values.

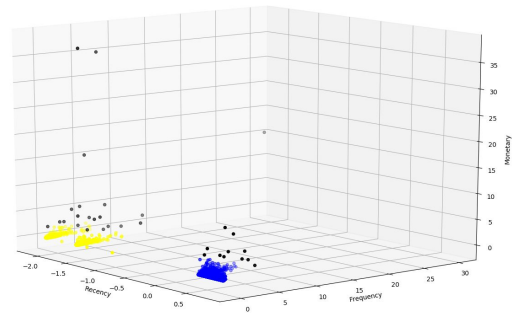

Fig. 6. Clusters calculated by using DBSCAN algorithm

As shown in Figure 6, DBSCAN algorithm obtained 2 clusters. These clusters are depicted with yellow and blue points. The black points represent outliers (noise) in a dataset. DBSCAN assigns these points a cluster number of -1. In this model we are mostly interested in outliers because, customer identified as an outlier either spends large amount of money, or makes transactions with extremely high frequency. Analyzing outliers gives us excessively valuable information, since in this dataset they represent the most valuable customers of Unibank. As it can be seen in Figure 6, outliers can further be clustered into 2 groups, based on the recency. Therefore it has been decided to use K-Means algorithm (with n_clusters = 2) on these outliers.

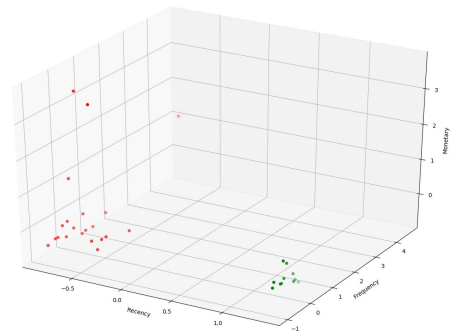

Fig. 7. Clusters calculated by K-Means on outliers

Figure 7. shows that K-Means, as expected, grouped outliers based on their recency. Later these clusters merged into the clusters obtained by the DBSCAN algorithm to get the final clusters acquired by using this model.

What is unique about this methodology is that if another model was used instead of the RFM to calculate the customer`s value it would be extremely challenging and sometimes nearly impossible to use DBSCAN algorithm to find these valuable clusters. This is the case, due to the 2 main drawbacks of DBSCAN which are determining parameter 'e' and finding clusters in varying densities in dimensions which are higher than 3. The first drawback exists because each time DBSCAN is applied the parameter 'e' has to be determined by using the algorithm above and even in 4 dimensional dataset it would be hard to visualize it hence estimating 'e' would become extremely challenging, since we won't be able to analyze it by making observations. The same issue happens when there are clusters with varying densities, since without observations it would be impossible to identify these clusters. However, since RFM is used, we only have 3 dimensions and we can always visualize the data and eliminate these 2 main drawbacks. These clusters can be identified by assigning small value of 'e' to find the densest cluster and then each time the value of 'e' can be increased to find the next clusters. After finding each cluster, it has to be extracted from the dataset so that at the end, after identifying all of the clusters the remaining data points will be outliers/noise.

### C. 3rd Model (Hierarchical Clustering)

Another clustering technique was used is Hierarchical clustering, which has two types:

1. Agglomerative.
2. Divisive.

In this model Agglomerative clustering was used, which calculates the predetermined number 'n' of the clusters by first considering each point in a dataset as a separate cluster then in each iteration it merges clusters together based on the euclidean distance until the predetermined number 'n' of the clusters obtained. Divisive clustering is just the opposite of the Agglomerative clustering.

In order to determine the number 'n' of the clusters dendrogram for the dataset was calculated.

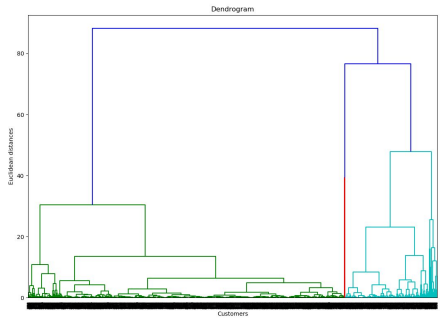

Fig. 8. Dendrogram

Based on the dendrogram shown in Figure 8, it has been decided to build a model using 4 clusters.

## V. DISCUSSION OF RESULTS

In this section the results obtained by applying the 3 models which were discussed in the implementation section are provided. To give more compact description of the results, for each of the models, first a figure illustrating all of the acquired clusters each with a unique color is depicted, then beneath that figure, the definition of the clusters are given. The obtained results have been verified by bank representatives and positively support our methodology.

### A. 1st Model(K-Means ➔ K-Means)

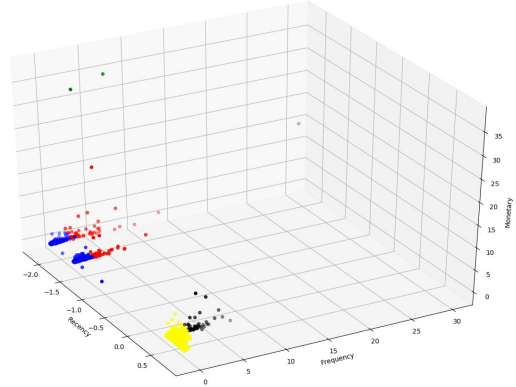

Fig. 9. Clusters obtained by applying 1st Model

**Black:** Customers who once were active by making frequent transactions but almost lost now.

**Yellow:** Customers who once were active by making few transactions and almost lost now.

**Red:** Customers who are active now and make transactions with a high frequency.

**Blue:** Customers who are active now but make few transactions.

**Green:** Customers who are active now, make frequent transactions and spend large amount of money.

### B. 2nd Model(DBSCAN ➔ K-Means)

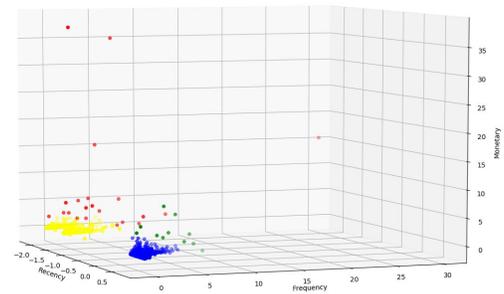

Fig. 10. Clusters obtained by applying 2nd Model

**Red:** Customers who are active now and spend tremendously large amount of money or make transactions with extremely high frequency or both.

**Green:** Customers who are inactive now and who once spent tremendously large amount of money or made transactions with extremely high frequency or both.

**Yellow:** Customers who are active now.

**Blue:** Customers who are inactive now.

*C. 3rd Model(Hierarchical Clustering)*

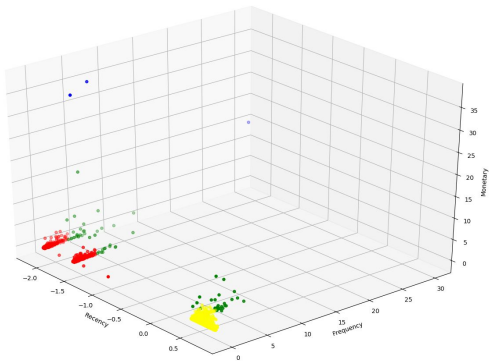

Fig. 11. Clusters obtained by applying 3rd Model

Hierarchical clustering produced similar clusters as K-means with 4 clusters and due to its high complexity problem, which is discussed in the comparison part it has been decided not to use it to build more sophisticated models.

*D. Comparison*

Hierarchical clustering is not efficient due to its computational complexity of $O(n^3)$, which means it cannot be used on large datasets. On the other hand, DBSCAN has computational complexity of $O(nlog(n))$ and it is crucial to find outliers in a dataset which represent most valuable customers of the financial institution. K-Means has computational complexity of $O(nkd)$ where, 'n' is the number of points, 'k' is the number of clusters, and 'd' is the number of attributes(features), which makes it the fastest among these three algorithms.

## VI. CONCLUSION

In our work, RFM model was used to calculate the customers value for the financial institution, then 3 models, which were based on 3 different types of clustering algorithms, were applied to the data, which contains RFM values of customers, to obtain customer segments.

In the first model K-Means algorithm were used twice to separate customers into 5 interesting clusters based on their recency, frequency and monetary backgrounds. The second model was built by the synthesis of DBSCAN and K-Means algorithm. First DBSCAN provided outliers/noise in the dataset, later K-Means was used on these outliers/noise to separate them into 2 clusters according to their recency. As a result, this model provided the most valuable customers for the financial institution. In the third model, agglomerative clustering algorithm with 4 clusters were applied to the dataset, which produced similar clusters as K-Means with 4 clusters. Therefore, for this reason and also due to its high computational complexity it has been decided to avoid this model.

As a future work, the methodology presented in this paper can be expanded, so that it can be applied to the other behaviours of bank customers, such as a behaviour based on deposits, loans, investments, etc. in addition to the ones that were extracted from the transactions history.